\documentclass[10pt, a4paper]{article}

\usepackage[final]{lrec2026} 

\usepackage{booktabs}

\usepackage[most]{tcolorbox}
\tcbset{
  cleanprompt/.style={
    colback=gray!3,
    colframe=black!20,
    boxrule=0.4pt,
    arc=2pt,
    fonttitle=\bfseries,
    coltitle=black,
    before skip=10pt, after skip=10pt,
    left=10pt, right=10pt, top=8pt, bottom=8pt,
    enhanced,
    breakable
  }
}

\title{From Slides to Chatbots: Enhancing Large Language Models with University Course Materials}

\name{Tu Anh Dinh*, Philipp Nicolas Schumacher*, Jan Niehues} 

\address{Karlsruhe Institute of Technology, Germany \\
         tu.dinh@kit.edu, philipp.schumacher@gmx.de, jan.niehues@kit.edu\\}

\abstract{
Large Language Models (LLMs) have advanced rapidly in recent years. One application of LLMs is to support student learning in educational settings. However, prior work has shown that LLMs still struggle to answer questions accurately within university-level computer science courses. In this work, we investigate how incorporating university course materials can enhance LLM performance in this setting. A key challenge lies in leveraging diverse course materials such as lecture slides and transcripts, which differ substantially from typical textual corpora: slides also contain visual elements like images and formulas, while transcripts contain spoken, less structured language. We compare two strategies, Retrieval-Augmented Generation (RAG) and Continual Pre-Training (CPT), to extend LLMs with course-specific knowledge. For lecture slides, we further explore a multi-modal RAG approach, where we present the retrieved content to the generator in image form. Our experiments reveal that, given the relatively small size of university course materials, RAG is more effective and efficient than CPT. Moreover, incorporating slides as images in the multi-modal setting significantly improves performance over text-only retrieval. These findings highlight practical strategies for developing AI assistants that better support learning and teaching, and we hope they inspire similar efforts in other educational contexts. 
 \\ \newline  }

\begin{document}

\maketitleabstract

\begingroup
\renewcommand\thefootnote{}
\footnotetext{*Co-first authors.}
\addtocounter{footnote}{-1}
\endgroup

\section{Introduction}

In recent years, the development of Large Language Models (LLMs) has advanced rapidly \cite{brown2020gpt3,grattafiori2024llama3,yang2024qwen2.5}. There has been growing interest in using LLMs to support teaching and learning, for example, by developing chatbots that can assist students in understanding complex course content. However, despite their general knowledge and reasoning abilities, current LLMs often struggle to provide accurate and contextually grounded answers to questions within university-level computer science courses. Benchmarks such as SciEx \cite{dinh2024sciex}, which evaluate models on real exam questions, show that while LLMs perform well in topics like Deep Learning and Neural Networks, they remain weak in others, such as Human–Computer Interaction.

One potential reason for these gaps is that LLMs may not have been exposed to the course-specific materials used in teaching. The materials, such as lecture slides and transcripts, tend to be different from the types of text LLMs are typically trained on. Although relatively small in scale, they contain highly specialized knowledge. Moreover, they have unique characteristics: slides include visual content such as images, formulas, and structured layouts, while transcripts contain spontaneous, conversational explanations. These properties make course materials both valuable and challenging sources for building educational chatbots.

In this work, we explore how incorporating course materials can improve LLM-based assistants in educational contexts. We focus on two strategies: (1) Retrieval-Augmented Generation (RAG), which dynamically retrieves relevant content from course materials to assist answer generation, and (2) Continual Pre-Training (CPT), which further trains the base model on the course materials. For lecture slides, we additionally investigate a multi-modal RAG approach, where visual information is preserved by presenting retrieved slides as images to the generator.

Our experiments, conducted on computer science course materials and evaluated on the SciEx benchmark, show that RAG is generally more effective and efficient than CPT, given the small dataset size of the course materials. Furthermore, leveraging slides in image form leads to notable gains over text-only retrieval. These results highlight practical strategies for constructing chatbots that better understand and communicate course-specific knowledge.

\section{Related Work} \label{sec:related_work}

\paragraph{Scientific Evaluation} Evaluating the capabilities of Large Language Models (LLMs) is crucial for understanding their potential as educational AI assistants. To evaluate the scientific capabilities of LLMs, several public benchmarks have been developed, such as SciQ \cite{welbl-etal-2017-SciQ}, ScienceQA \cite{lu2022ScienceQA}, and M3Exam \cite{zhang2024m3exam}. However, the questions in these benchmarks are multiple-choice, which do not reflect the open-ended interactions expected from an educational assistant. In real learning scenarios, students are likely to prefer free-form, natural-language explanations, rather than simple answer selection.
The SciEx benchmark \cite{dinh2024sciex} addresses this limitation by evaluating LLMs on university-level computer science exam questions, including a broader range of question types beyond multiple-choice. SciEx has shown that LLMs still underperform in certain topics. Since each SciEx exam corresponds to a real university course, the associated course materials, i.e., lecture slides and transcripts, offer a promising resource for improving the LLMs. In this work, we leverage such materials to enhance LLMs’ ability to function as educational assistants, and use SciEx as our evaluation benchmark.

\paragraph{Retrieval-Augmented Generation}
Retrieval-Augmented Generation (RAG) \cite{rag_introduction_2020_meta} combines external knowledge retrieval with generative language models to provide informed responses. Recent work has focused on adapting RAG to scientifics domains. For example, \citet{gokdemir2025hiperrag} proposed HiPerRAG, which scales retrieval to millions of scientific articles. \citet{shi2025hypercube} introduced Hypercube-RAG, a multidimensional retrieval method that better captures the structure of scientific knowledge. Domain-specific RAG benchmarks such as ChemRAG-Bench \cite{zhong2025benchmarking} demonstrate substantial gains in performance on chemistry-related tasks. Beyond text retrieval, multimodal RAG systems such as \citet{matsumoto2024kragen} integrate knowledge graphs to enhance reasoning in biomedical contexts. However, most of these works rely on large-scale, text-only corpora, differing to our work, which investigates RAG using the small, specialized and partially multimodal source of university course materials.

\paragraph{Continual Pre-Training for Domain Adaptation}
Continual Pre-Training (CPT) is an effective method for domain adaptation by continuing the pre-training of a general LLM on domain-specific data, as shown by \citet{gururangan-etal-2020-dont}. Subsequent studies have examined strategies to optimize CPT and mitigate catastrophic forgetting. \citet{que2024d} proposed the D-CPT Law to balance the mixture of domain and general corpora. \citet{yildiz2024investigating} analyzed learning dynamics during CPT, indicating the importance of monitoring domain and general performance for early stopping. \citet{ibrahim2024scalable} showed that a good selection of learning rate re-warming, learning rate re-decaying, and replay of previous data could be sufficient to match the performance of fully re-training from scratch. CPT has been successfully applied in domains such as biomedical \cite{jin2023medcpt} and financial \cite{xie2024efficientcpt}. Differing to previous works, our work investigates CPT on very limited but specialized data from university course materials, which poses additional risks of overfitting and forgetting.

An important consideration is how and at which stage of model training CPT should be performed to preserve the model’s instruction-following capability. \citet{jindal2024balancingcpt} compared two strategies: (1) applying continual pre-training directly on an instruction-tuned model, and (2) continually pre-training the base model followed by instruction fine-tuning. Their results show that the first approach leads to catastrophic forgetting of instruction-following abilities, whereas the second approach better preserves them. However, instruction fine-tuning is computationally expensive, requiring large amounts of hand-annotated instruction data. Therefore, the authors proposed an alternative: adding \textit{instruction residuals} to the base LLM after CPT to restore the instruction-following capability. 
We employ this approach in our work to avoid the expensive instruction fine-tuning. 

\section{Gathering Course Materials Data} \label{sec:course_materials}

We obtain course materials corresponding to the university exams used in the SciEx benchmark \cite{dinh2024sciex} by contacting the dataset creators. These materials provide an authentic educational context from which the exam questions were drawn, allowing us to more realistically evaluate LLMs as potential assistants for university-level education.

We focus on two types of materials: lecture slides and lecture transcripts. These resources are widely available in most university courses and thus represent a practical solution for building educational chatbots. At the same time, they differ substantially in structure and modality. Slides are concise, visually organized, and often include non-textual information such as images and formulas. Transcripts, in contrast, contains spoken language, offering more explanations and contextual details. We aim to explore how LLMs can leverage diverse course material formats to better understand and generate course-specific explanations.


\paragraph{Slides}
We collected the slides as PDF files, where each PDF page corresponds to a separate slide in the presentation. 
For experiments with text-only LLMs, we extract the text from the PDFs. For experiments with visual-aided LLMs, we convert the PDF pages to images.

\begin{figure*}[h]
    \centering
    \includegraphics[width=0.8\linewidth]{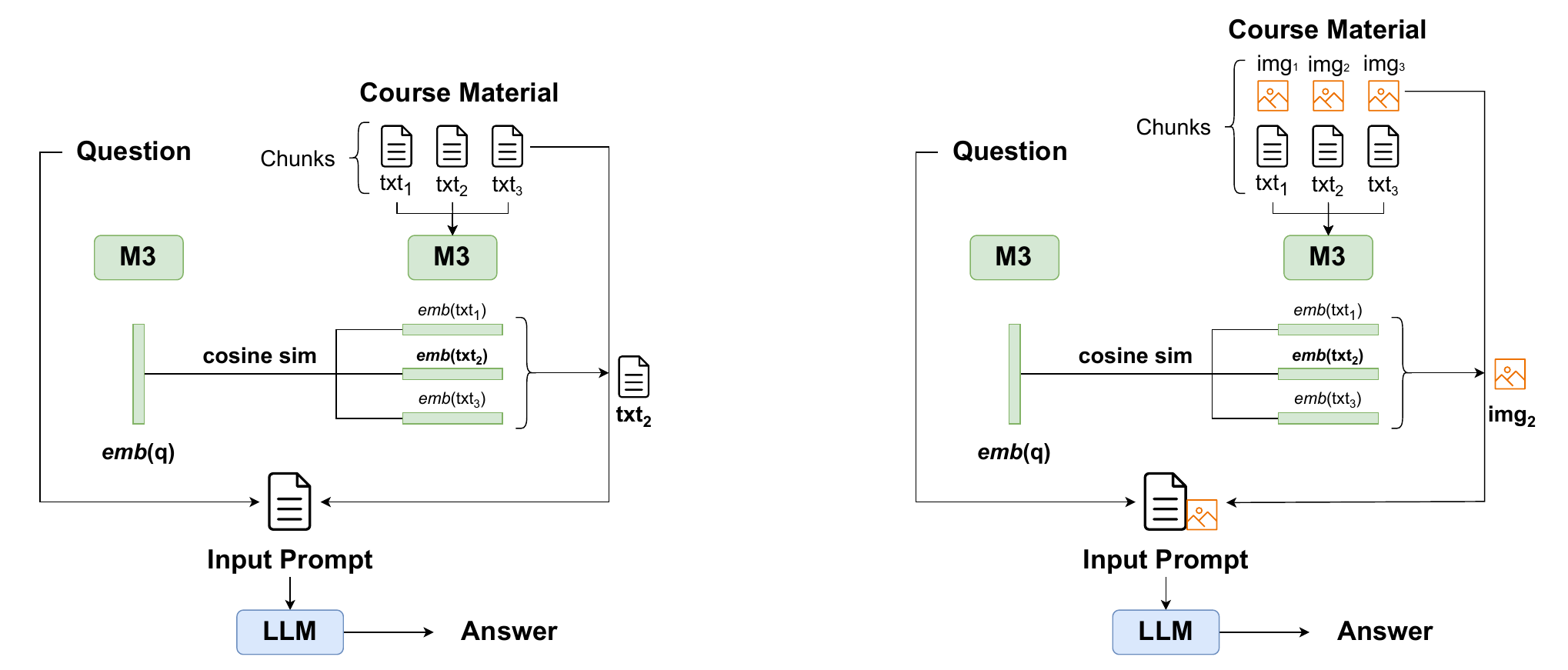}
    \caption[Course Material Incorporation: RAG Workflow]{
        \textbf{Retrieval-Augmented Generation (RAG) workflow.}
        The retriever encodes slide text and lecture transcripts into a text-based vector database. 
        At inference time, relevant text chunks or slides are retrieved. Left-hand side: \textbf{text-only RAG}. Right-hand side: \textbf{multimodal RAG}:
        when using a visually capable LLM, retrieved slides are passed as \emph{images} to the generator, preserving visual information.
    }
    \label{fig:own_rag}
\end{figure*}

\paragraph{Transcripts}
We extract the textual transcriptions from the audios of the lecture recordings using the Lecture Translator framework \cite{huber-etal-2023-end}, which employs \textit{Whisper Large V2} \cite{radford2023whisper} paired with SHAS segmentation \cite{tsiamas2022shas}. We clean up the transcriptions afterward by manually checking and removing hallucinated sentences, i.e., sentences repeatedly presented in the transcriptions but not presented in the audio.

As the transcripts are from spoken language, they are non-fluent, which might be unsuitable for direct usage with LLMs. Therefore, we polish the transcriptions using the 4-bit-quantized Qwen2.5-Instruct 72B model.
We ask the model to make the text more fluent and remove irrelevant content.
We split the transcriptions into chunks of five sentences, and include the previously generated polished chunk in each input prompt to ensure smooth transitions.
Finally, the resulting polished transcripts are manually processed to remove LLM-generated sentences that are irrelevant to the actual transcript. 
The detailed prompt is in Appendix \ref{app:prompt_polish_transcript}.

\paragraph{Data Statistics}  
The final dataset includes course materials in the form of lecture slides and transcriptions in English and German. The dataset covers six topics:  Algorithms (ALGO), Databases (DBS), Deep Learning and Neural Networks (DLNN), Human–Computer Interaction (HCI), Natural Language Processing (NLP), and Theoretical Foundations of Computer Science (TGI). Note that for DBS, there are no transcriptions available. In total, the slide materials comprise 3.9K slides containing 0.2M tokens, with an average of approximately 50 tokens per slide. The lecture transcriptions consist of 3.3M tokens.

\section{Incoorporating Course Material}
We extend LLMs' knowledge with the course materials in two different ways: Retrieval-Augmented Generation (RAG) and Continual Pre-Training (CPT).

\subsection{Retrieval-Augmented Generation}

In Retrieval-Augmented Generation (RAG), the goal is to retrieve relevant information from the course materials and include it in the input to the LLM at inference time. The overall workflow is shown in Figure~\ref{fig:own_rag}.

\paragraph{RAG Components} 
We follow the standard RAG architecture, which consists of a \emph{retriever} and a \emph{generator}. The retriever identifies the most relevant segments of course material, while the generator (the LLM) uses these segments to produce an answer. The generator is the model evaluated for its ability to serve as an educational assistant.  
For retrieval, we adopt the pre-trained \textbf{M3-Embedding} model \cite{multi2024m3}, chosen for its multilingual coverage (as our materials are in both English and German) and its compact size (559M parameters), which makes it efficient for practical use.

\begin{figure*}[h]
    \centering
    \includegraphics[width=0.65\linewidth]{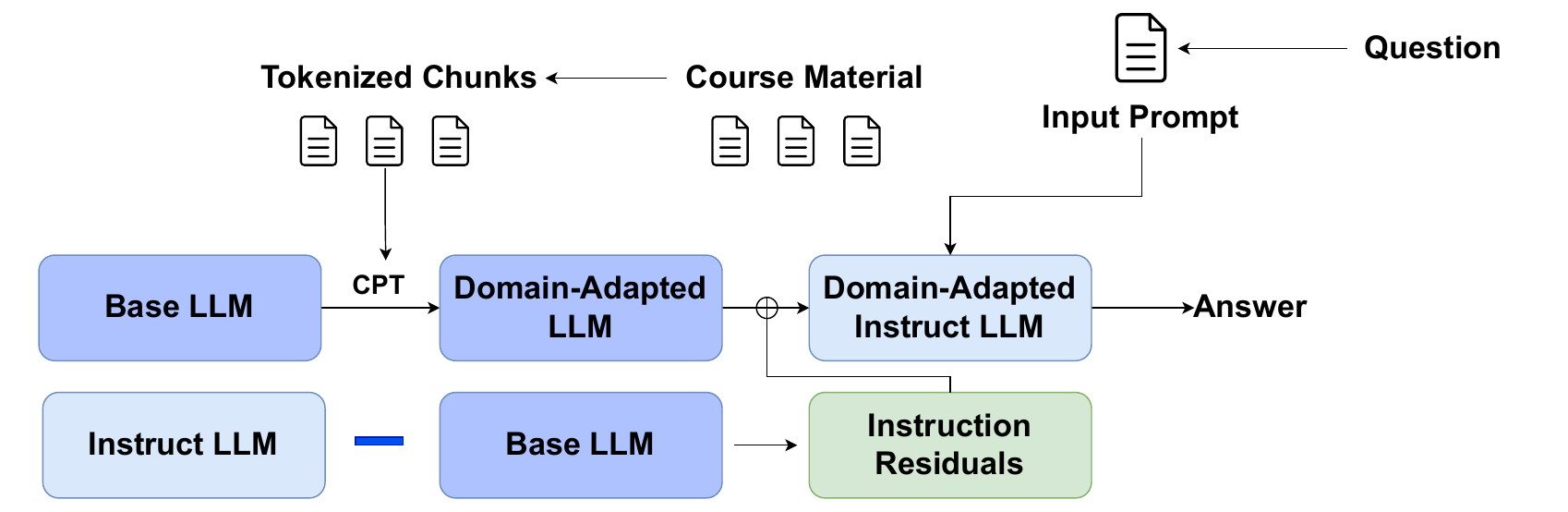}
    \caption[Course Material Incorporation: CPT Workflow]{
        CPT workflow, assuming the availability of both base and instruction-tuned versions of the LLM.
    }
    \label{fig:own_cpt}
\end{figure*}

\paragraph{RAG Knowledge Base} 
To build the knowledge base, we segment course materials into semantically coherent chunks, each ideally representing an independent concept. For slides, each slide is treated as one chunk. For lecture transcripts, which are continuous in nature, we segment the text sequentially while ensuring that sentences remain intact and that each chunk does not exceed a pre-defined maximum size. We also apply a 10\% overlap between consecutive chunks to reduce context loss. Each chunk is then embedded using the M3 model to create the vector database.

\paragraph{RAG Workflow} 
At inference time, a student question is embedded using the same M3 model as the course materials. Cosine similarity is then used to retrieve the $k$ most relevant chunks from the vector database. These retrieved chunks are passed to the generator as contextual information for answer generation. The question and the retrieved content are clearly separated in the prompt to help the model distinguish between them, and the model is instructed to rely on the retrieved content only when relevant.

\paragraph{Multi-Modal RAG for Slide Materials} 
A unique aspect of our approach is a \textbf{multi-modal RAG setup} designed for \emph{visually capable LLMs}. While the retriever always operates on text embeddings, obtained from the textual content extracted from slides, the generator receives the \emph{actual slide images} instead of their text representation. This design preserves visual information that would otherwise be lost during text extraction, such as images, slide layout, and formulas. In this configuration, the retriever efficiently locates relevant slides via text embeddings, and the generator benefits from the richer visual context during response generation.

\subsection{Continual Pre-Training} 
In Continual Pre-Training (CPT), we continue pre-training the base LLMs on the next word prediction task using course material data (Figure \ref{fig:own_cpt}). 

\paragraph{Data Preprocessing} The text is divided into semantically logical chunks, following the same procedure as in RAG to respect the LLM’s maximum context length. However, unlike for RAG, we do not maintain overlaps between chunks.

\paragraph{Continual Pre-Training} Special care is required when performing CPT with the very limited course material data, as the model is otherwise prone to catastrophic forgetting of previously acquired knowledge. To mitigate this risk, we reduce the maximum learning rate (LR) relative to the original pre-training process and apply both LR warm-up and LR decay. In addition, we replay a portion of the original pre-training data during CPT. Validation perplexity on both the original pre-training data and the course material data is monitored to determine an appropriate early stopping point, preventing overfitting to the course material.

\paragraph{Instruction Following} After CPT, we aim to restore instruction-following capabilities of the LLM. To achieve this, we adopt the approach of \citet{jindal2024balancingcpt}, which introduces Instruction Residuals (IRs). This method leverages a pre-existing instruction-tuned version of the base LLM on which CPT was performed. The IRs are computed by subtracting the weights of the base model, denoted as $\theta_b$, from those of the instruction-tuned model, denoted as $\theta_i$:

\begin{equation}
    IR(\theta_i, \theta_b) = \theta_i - \theta_b
\end{equation}

Then, the residuals $IR(\theta_i, \theta_b)$ are added to restore the instruction-following of a continually pre-trained base LLM with weights $\theta_{b_{cpt}}$:

\begin{equation}
    \theta_{i_{cpt}} = \theta_{b_{cpt}} + IR(\theta_i, \theta_b)
\end{equation}

The resulting model with weights $\theta_{i_{cpt}}$ is expected to be adapted with the course materials while retaining its ability to follow instructions.

\begin{table*}[ht]
\small
\centering
    \begin{tabular}{l l c c c}
    \toprule
    \textbf{} & \textbf{Full name} & \textbf{\# Params} & \textbf{Quant.} & \textbf{Handle Image} \\
    \midrule
    LLaMA 3.1 & meta-llama/Llama-3.1-8B-Instruct & 8B & - & no \\
    LLaMA 3.3 & meta-llama/Llama-3.3-70B-Instruct & 70B & 4 bit & no \\
    Qwen2-VL & Qwen/Qwen2-VL-72B-Instruct-GPTQ-Int4 & 72B & 4 bit & yes \\
    Qwen2.5 & Qwen/Qwen2.5-72B-Instruct & 72B & 4 bit & no \\
    Mistral Large 2 & mistralai/Mistral-Large-Instruct-2411 & 123B & 4 bit & no \\
    \bottomrule
    \end{tabular}
\caption[Experimental Setup: LLMs Used]{
        Details of the LLMs used.
    }
\label{tab:llm_details}
\end{table*}

\section{Experimental Setup} 

\subsection{Data} \label{sec:data}
\paragraph{Course Materials} For extending LLMs' knowledge, we use the course materials data detailed in Section \ref{sec:course_materials}.

\paragraph{Replayed Data for CPT} As replayed data to be used with CPT, we use English and German Wikipedia data. The amount of Wikipedia data used is half the amount (in terms of tokens) of the course materials data, as we defined the replay ratio to be 33\% (detailed in Section \ref{sec:Hyperparameters}).

\paragraph{Evaluation Data} As discussed in Section \ref{sec:related_work}, we use the SciEx benchmarking dataset \cite{dinh2024sciex} to evaluate our approaches.  SciEx has in total 10 exams and  154 unique questions. Each question is annotated with a difficulty level among \{\textit{easy, difficult, hard}\}. SciEx also comes with answers to the questions from different LLMs, along with expert-annotated scores on the answers. The provided expert scores on pre-generated answers enable us to evaluate potential LLM graders for automatic evaluation used in our paper.

\subsection{Models} \label{sec:models}
The list of all LLMs used in our experiments can be found in \autoref{tab:llm_details}.
All of them are instruction-tuned versions of the base models and available on Hugging Face\footnote{\url{https://huggingface.co/docs/hub/en/models-the-hub}}.

\paragraph{Models For Grading}
LLaMA 3.3 70B and Qwen2.5 72B are used as candidate grader LLMs for evaluation, as they are recent models with large parameter sizes. We additionaly include  Mistral Large 2 123B as a candidate grader, as LLaMA 3.3 70B and Qwen2.5 72B might exhibit biases when grading their own answers. 

\paragraph{Models For RAG}
We experiment with LLaMA 3.3 70B, Qwen2.5 72B, LLaMA 3.1 8B, and Qwen2-VL 72B in the RAG setting. LLaMA 3.3 and Qwen2.5 are included as they represent the flagship models of their respective families at the time of the experiments. LLaMA 3.1 8B is evaluated to examine whether incorporating course materials has a different effect on older, smaller models. Finally, Qwen2-VL is tested to assess the impact of including course material images, as it can directly process visual inputs.

\paragraph{Models For CPT}

We experiment with LLaMA 3.1 8B for Continual Pre-Training (CPT), as its smaller size makes CPT more computationally feasible. Using LLaMA 3.1 8B also enables a direct comparison between CPT and RAG, since the same model is employed in both settings.

\subsection{Hyperparameters} \label{sec:Hyperparameters}
\paragraph{RAG} For RAG, we experiment with different numbers of retrieved chunks $k$. For slide chunks, we vary $k$ between 1 and 20 and find that $k=4$ yields the best average performance. For transcripts, we fix $k=4$ and vary the chunk size across 100, 300, and 750 tokens, observing that 300 tokens provides the best results.

\paragraph{CPT} We use an effective batch size of 32, a maximum sequence length of 2048, a cosine learning rate schedule, and a weight decay of 0.1. Since CPT is performed on a very small dataset, careful tuning of the learning rate is crucial. We conduct a grid search with CPT on synthetic transcripts training data from the Human–Computer Interaction (HCI) lecture, varying the learning rate from 3e-6 to 6e-5 and the number of warm-up steps among {0, 5, 10}. The best configuration is a learning rate of 2e-5 with 5 warm-up steps, which is significantly lower than the original maximum pre-training learning rate of 3e-4 for LLaMA 3.1. We then apply these hyperparameters to the final setting, in which we combine the slides and the synthetic transcripts of all courses for CPT, given the limited dataset size.
The training employs a replay ratio of 33\%, i.e., 33\% of the training tokens is Wikipedia data. CPT is run for a maximum of 15 epochs, but training is stopped early once validation domain perplexity ceases to improve. We employ training-to-validation split ratio of 90:10.

\subsection{Evaluation}
We employ automatic grading with an LLM to evaluate the proposed approaches on the university exam questions in SciEx. As candidate graders, we consider LLaMA 3.3, Qwen2.5, and Mistral Large 2. Among these, Mistral Large 2 performs best, achieving a Pearson correlation of 0.84 with the expert-provided grades in the SciEx dataset.

Adhering to SciEx, the exam-level scores range from 0 to 60 points. Question-level scores have different ranges, so we report the normalized scale, ranging from 0 to 100.

In all experiments using LLMs to solve questions in SciEx, we use vanilla prompting without incorporating course materials as the baseline for the knowledge-enhanced approaches (RAG and CPT).

The prompt used to generate answers is in Appendix \ref{app:prompt_answer}.  The prompt used to grade exams is in Appendix \ref{app:prompt_grade}.

\subsection{Hardwares}
Inferencing is performed on either an NVIDIA A100 (80 GB), an H100 (94 GB), or two V100s (32 GB). Training is performed on four NVIDIA H100s.

\section{Results and Discussion}

\subsection{LLM Performance with RAG}

To understand the effectiveness of Retrieval-Augmented Generation (RAG) in the educational context, we analyze how different factors influence its performance. We examine model size and architecture, course topics, question difficulty, the modality and type of course materials used.

\subsubsection{Effect of Model Architecture and Size}
\paragraph{RAG consistently improves model performance} Table \ref{tab:llm_solver_diff} shows the performance of each LLM with and without RAG using slide text. Overall, incorporating RAG with slide text improves the performance of all models. Among them, LLaMA 3.1 8B, the smallest model, shows the least improvement (+0.8 points), likely due to its more limited capacity to leverage additional context.

\begin{table}[ht]
\centering
\small
\begin{tabular}{lcc}
\toprule
Model & Baseline & RAG Slide Text \\
\midrule
Qwen2.5   & 45.27 & \textbf{47.36}  \\
LLaMA 3.3 & 43.10 & \textbf{44.01} \\
Qwen2-VL  & 38.05 & \textbf{40.87} \\
LLaMA 3.1 & 31.17 & \textbf{31.97} \\
\midrule 
Average & 39.40 & \textbf{41.05} \\
\bottomrule 
\end{tabular}
\vspace{0.3em}
\caption[RAG Exam Solving with Slide Text: Average Exam Performance per LLM]{%
    RAG with slide text: average performance across all exams for each LLM. (Exam-level scores, 0-60 range.)
}
\label{tab:llm_solver_diff}
\end{table}

\subsubsection{Effect of Course Topics}
\paragraph{RAG mitigates knowledge gaps} 
Table \ref{tab:llm_exam_comparison} presents the average performance across exams from different topics. On certain topics, such as Human–Computer Interaction (HCI), RAG yields substantial improvements (+6.11 points). For others, performance either improves modestly or remains stable. As \citet{dinh2024sciex} observed, LLMs often lack context for HCI, which our results suggest can be partially addressed by retrieving course-specific materials. RAG thus appears to be the most useful for filling knowledge gaps while maintaining performance in areas where LLMs are already strong.

\begin{table}[ht]
\centering
\small
\begin{tabular}{lcc}
\toprule
Exam & Baseline & RAG Slide Text \\
\midrule
HCI\textsuperscript{1}   & 41.91 & \textbf{48.02}  \\
DBS\textsuperscript{2}   & 30.16 & \textbf{32.56}  \\
NLP\textsuperscript{1}   & 43.76 & \textbf{44.34}  \\
Algo  & \textbf{35.38} & 35.13  \\
TGI   & \textbf{41.38} & 41.25  \\
DLNN\textsuperscript{1}  & \textbf{42.78} & 42.15  \\
\midrule 
Average & 39.40 & \textbf{41.05}  \\
\bottomrule
\end{tabular}
\vspace{0.3em}
\begin{minipage}{0.8\columnwidth}
\small
\textsuperscript{1} Averaged over English and German Exams \\
\textsuperscript{2} Averaged over 2022 and 2023 Exams \\
\end{minipage}
\caption[RAG Exam Solving with Slide Text: Average LLM Performance per Exam]{%
    RAG with slide text: average performance across all LLMs for each exam topic. (Exam-level scores, 0-60 range.)
}
\label{tab:llm_exam_comparison}
\end{table}

\subsubsection{Effect of Question Difficulty}
\paragraph{RAG provides greater benefits on difficult questions} 
Table \ref{tab:rag_difficulty} reports the impact of RAG when questions are grouped by difficulty levels (as defined in SciEx). RAG yields the largest gains on hard questions (+10.79\%). In contrast, improvements on easy and medium questions are smaller.

\begin{table}[ht]
\centering
\small
\begin{tabular}{lccr}
\toprule
Difficulty & Baseline & RAG Slide Text & $\Delta$* \\
\midrule
Easy   & 71.71 & 73.63 & 1.92  \\
Medium & 66.83 & 72.16 & 5.33 \\
Hard  & 56.72 & 67.51 & 10.79 \\
\bottomrule 
\end{tabular}
\vspace{0.3em}
\centering
\begin{minipage}{0.9\columnwidth}
\small
* RAG improvement over the baseline.
\end{minipage}
\caption[RAG Exam Solving with Slide Text: Average Performance per difficulty level of questions]{%
    RAG with slide text: average performance across questions of different difficulty levels. (Question-level scores, range 0-100.)
}
\label{tab:rag_difficulty}
\end{table}


\subsubsection{Effectiveness of Multimodal RAG}
\begin{figure*}[h]
    \centering
    \includegraphics[width=0.99\linewidth]{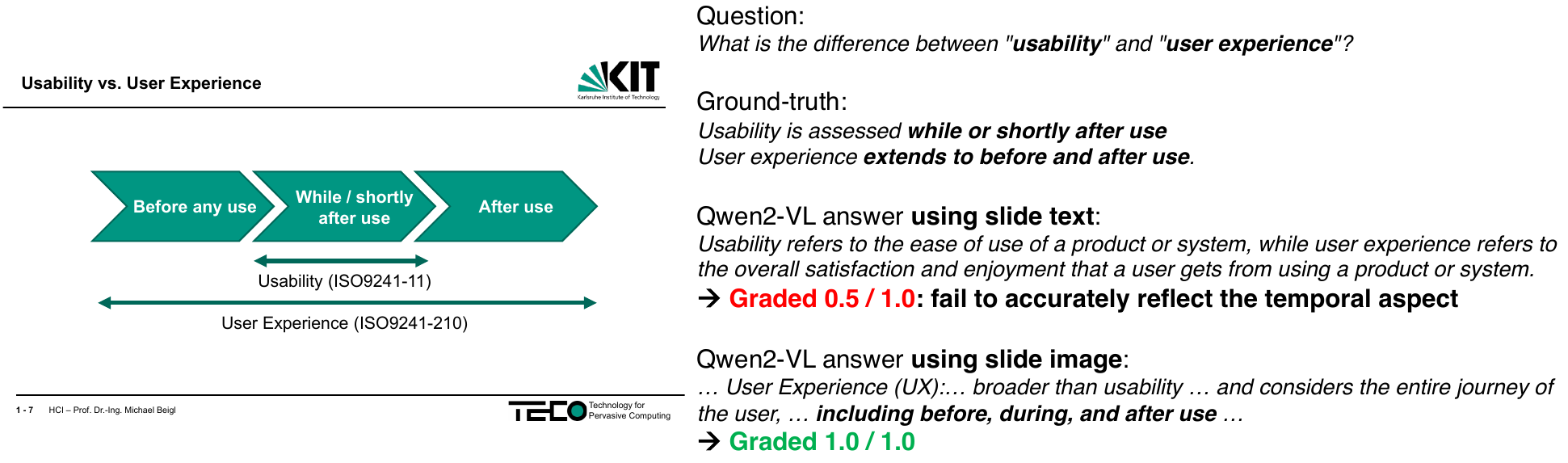}
    \caption{Example of multimodal RAG effectiveness: Qwen2-VL produces a more accurate answer when provided with the slide as an image rather than as plain text. Viewing the slide image preserves the visual context, specifically, \textbf{the arrows illustrating the temporal scope of \textit{usability} and \textit{user experience}}.}
    \label{fig:example_slide_image}
\end{figure*}

\paragraph{Slides as images outperform text extraction} 
We examine whether visually capable LLMs can benefit from the non-textual aspects of slides under our multimodel RAG setting. Using Qwen2-VL, we compare slide text retrieval versus slide image input (Table \ref{tab:image_context}). The results show that providing slide images leads to higher performance, confirming that visual information, such as images, layout, and formulas, contributes to better comprehension. Figure \ref{fig:example_slide_image} shows an example of the effectiveness of multimodal RAG, where Qwen2-VL gives better answers by leveraging the visual information in the retrieved slide.

\begin{table}[ht]
\centering
\small
\begin{tabular}{lccc}
\toprule
Exam & Base Model & RAG & RAG \\
  & & slide text & slide image \\
\midrule
HCI\textsuperscript{1}   & 39.60 & 47.20 & \textbf{50.75}  \\
DBS\textsuperscript{2}   & 23.03 & 26.15 & \textbf{32.88}  \\
NLP\textsuperscript{1}   & 44.62 & 43.88 & \textbf{44.50}  \\
Algo  & 38.50 & \textbf{42.00} & 37.00  \\
TGI   & 43.50 & 41.50 & \textbf{48.00}  \\
DLNN\textsuperscript{1}  & 41.98 & \textbf{45.38} & 41.85  \\
\midrule 
Average & 38.05 & 40.87 & \textbf{42.50}  \\
\bottomrule
\end{tabular}
\vspace{0.3em}
\begin{minipage}{0.94\columnwidth}
\small
\textsuperscript{1} Averaged over English and German Exams \\
\textsuperscript{2} Averaged over 2022 and 2023 Exams \\
\end{minipage}
\caption[RAG Exam Solving with Slide Images: Comparison to Baselines]{%
    RAG slide text versus slide image: effect on Qwen2-VL performance. (Exam-level scores, range 0-60.)
}
\label{tab:image_context}
\end{table}

\subsubsection{Effect of Course Material Types}
\paragraph{Slides outperform lecture transcriptions for RAG} Table \ref{tab:synth_transcripts_exam} compares the effectiveness of using lecture transcriptions (both raw and LLM-polished) versus slides as materials for RAG. Unlike slides, transcriptions fail to provide improvements and, in many cases, even reduce performance compared to the baseline. A likely explanation is that slides contain concise and well-structured text, whereas transcriptions consist of spoken language which tends to be more more verbose and less organized, thus may distract the model from essential information. In the following experiments, when comparing CPT to RAG, we report the performance of RAG with slide text.

\begin{table}[ht!]
    \centering
    \small
    \setlength{\tabcolsep}{4pt}
    \begin{tabular}{l c c c c}
        \toprule
        & Baseline & Slide & Transcripts & Polished \\
        & & Text & & Transcripts \\
        \midrule
        HCI\textsuperscript{1} & 41.91 & \textbf{48.02} & 40.13 & 40.13 \\
        TGI & 41.38 & 41.25 & \textbf{41.38} & 39.75 \\
        DLNN\textsuperscript{1} & \textbf{42.78} & 42.15 & 42.05 & 39.12 \\
        NLP\textsuperscript{1} & 43.76 & \textbf{44.34} & 42.89 & 42.96 \\
        Algo & \textbf{35.38} & 35.13 & 34.25 & 32.50 \\
        \hline 
        Average & 41.70 & \textbf{43.17} & 40.72 & 39.58 \\
        \bottomrule
    \end{tabular}
    
    \begin{minipage}{0.95\columnwidth}
        \small
        \textsuperscript{1} Averaged over English and German Exams \\
    \end{minipage}
    \caption[RAG Exam Solving with Synth. Transcripts: Comparison to Baselines]{%
        RAG slide text versus transcript: Effect on performance on each topic. DBS is excluded as there are no transcripts. (Exam-level scores, range 0-60.)
    }
    \label{tab:synth_transcripts_exam}
\end{table}

\subsection{LLM Performance with CPT}

We analyze how CPT on course materials effect LLaMA~3.1 8B performance. We examine the model's learning dynamics, overall performance, and sensitivity to different types of course materials. In the experiments, we perform CPT using combined slide text and polished transcripts from all courses, unless stated otherwise.

\subsubsection{Adaptation Dynamics During CPT}
\begin{figure}[ht]
    \centering
    \includegraphics[width=\linewidth]{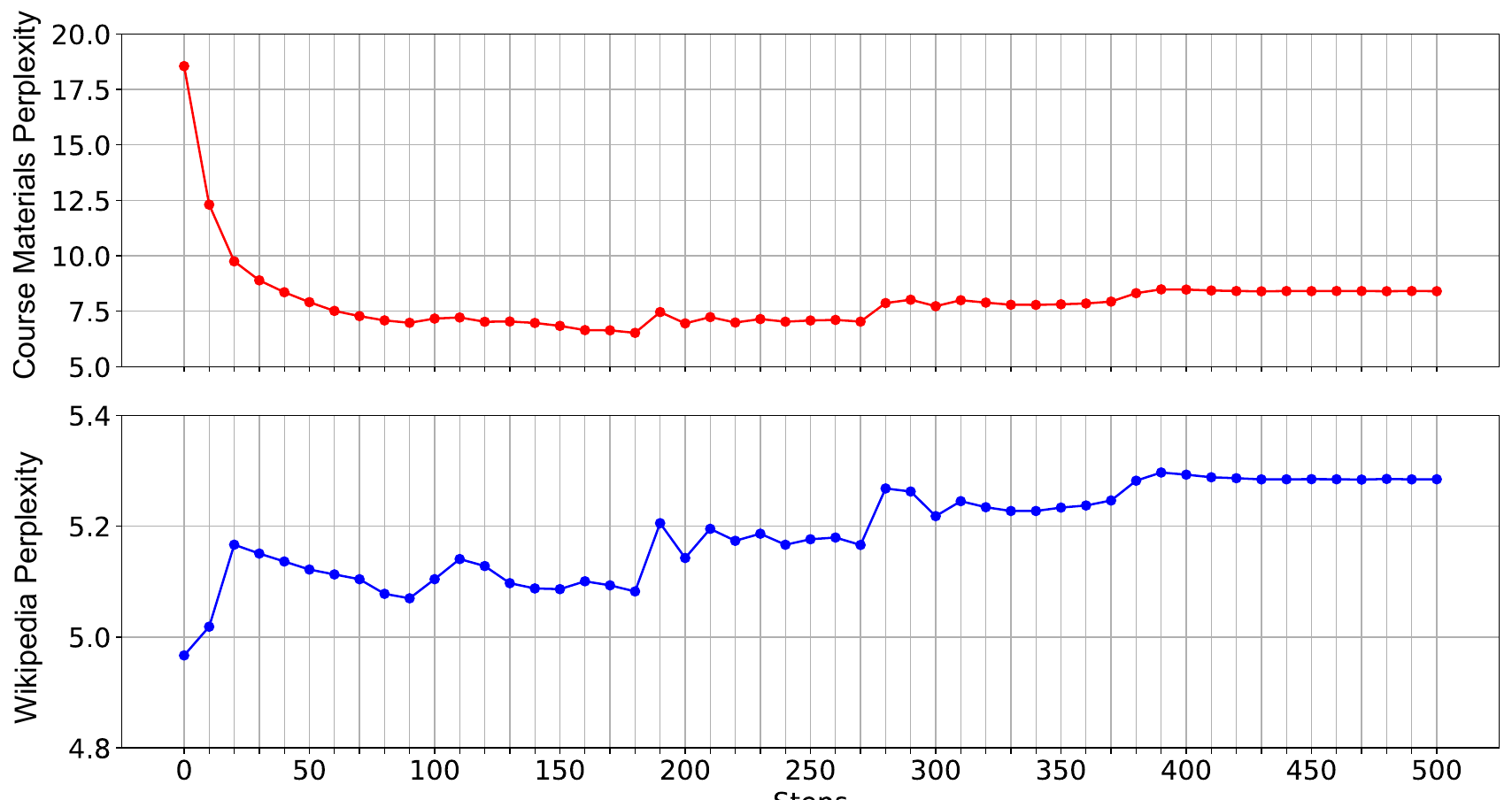}
    \caption[CPT with All Lectures Combined Materials: Perplexities]{
        Course materials and Wikipedia (replay) validation perplexity for CPT with LLaMA 3.1 8B.
    }
    \label{fig:05_cpt_all_lecture_slides_and_synthetic_transcripts}
\end{figure}

\paragraph{LLaMA 3.1 8B can learn from course materials} Figure \ref{fig:05_cpt_all_lecture_slides_and_synthetic_transcripts} shows the validation losses (measured in perplexity) during CPT on course material data and Wikipedia data. The validation perplexity on the course materials drops sharply in the first few steps before stabilizing, whereas the Wikipedia validation perplexity gradually increases over time. By applying early stopping once the course material perplexity converges, we obtain a model that adapts effectively to the course materials while minimizing forgetting of its original knowledge.

\subsubsection{Overall Performance}
\paragraph{CPT does not bring consistent improvement} Table \ref{tab:cpt_all_exams_results} compares the performance of CPT with the baseline and RAG. On average, CPT performs worse than both the baseline and RAG. To further investigate, we break down the results by exam language, focusing on exams available in both English and German (Table \ref{tab:llm_language_effect_cpt}). The results show that CPT yields improvements over the baseline in some cases (e.g., the HCI English exam, the NLP German exam, and the DLNN German exam), but causes significant drops in others (notably the HCI German exam). The overall loss in performance is likely due to catastrophic forgetting, which is difficult to avoid when conducting CPT on the limited amount of course material data.

\begin{table}[ht!]
    \centering
    \small
    \begin{tabular}{l c c c}
        \toprule
        \multicolumn{4}{c}{\textbf{LLaMA 3.1}} \\
        \midrule
        & \textbf{Baseline} & \textbf{RAG} & \textbf{CPT} \\
        \midrule
        HCI\textsuperscript{1} & 39.40 & \textbf{42.85} & 37.85 \\
        DBS\textsuperscript{2} & 21.93 & \textbf{26.35} & 21.85 \\
        TGI & \textbf{29.00} & 26.00 & 18.00 \\
        DLNN\textsuperscript{1} & \textbf{34.12} & 29.62 & 31.88 \\
        NLP\textsuperscript{1} & 34.75 & \textbf{37.25} & 34.25 \\
        Algo & \textbf{23.00} & 22.00 & 19.00 \\
        All Exams & 31.17 & \textbf{31.97} & 28.86 \\
        \bottomrule
    \end{tabular}
    \begin{minipage}{0.94\textwidth}
        \small
        \textsuperscript{1} Averaged over English and German Exams \\
        \textsuperscript{2} Averaged over 2022 and 2023 Exams \\
    \end{minipage}
    \caption[Cross-Course CPT Exam Solving: Performance on All Exams]{%
        CPT: LLaMA 3.1's performance across exam topics. (Exam-level scores, range 0-60.)
    }
    \label{tab:cpt_all_exams_results}
\end{table}

\begin{table}[ht!]
  \centering
  \small
    \begin{tabular}{lccc | c}
    \toprule
    \multicolumn{5}{c}{\textbf{LLaMA 3.1}} \\
    \midrule
      
        & \textbf{HCI\textsuperscript{1}}
        & \textbf{NLP\textsuperscript{2}}
        & \textbf{DLNN\textsuperscript{2}}
        & \textbf{All Exams} \\
      \midrule
      \multicolumn{5}{l}{\textbf{English}} \\
      \cmidrule(lr){1-5}
      Baseline  & 38.70 & 37.00 & \textbf{34.30} & 36.67 \\
      RAG   & \textbf{42.40} & \textbf{42.00} & 31.75 & \textbf{38.72} \\
      CPT  & 41.90 & 35.50 & 31.68 & 36.36 \\

      \addlinespace[0.5em]
      \multicolumn{5}{l}{\textbf{German}} \\
      \cmidrule(lr){1-5}
      Baseline   & 39.40 & 32.50 & \textbf{33.95} & \textbf{35.28} \\
      RAG   & \textbf{42.85} & 32.50 & 27.50 & 34.28 \\
      CPT  & 33.80 & \textbf{33.00} & 32.08 & 32.96 \\
      \bottomrule
    \end{tabular}
    
    \begin{minipage}{0.67\textwidth}
        \small
        \textsuperscript{1} Provided course material was English and German \\
        \textsuperscript{2} Provided course material was English \\
    \end{minipage}
  \caption[Cross-Course CPT Exam Solving: Impact of Exam Language]{%
        CPT breakdown by language: LLaMA 3.1's performance on German and English exams. (Exam-level scores, range 0-60.)
  }
  \label{tab:llm_language_effect_cpt}
\end{table}

\subsubsection{Effect of Course Material Type}
\paragraph{Polished transcripts yield the best CPT performancee} We conduct an ablation study to examine the effect of different types of course materials for CPT on model performance. Due to computational constraints, this experiment is limited to the Human–Computer Interaction (HCI) topic. The results are shown in Table \ref{tab:cpt_hci_results}. We observe the same pattern: CPT in general performs worse than RAG. Among the CPT configurations, the best results are achieved when using polished transcripts. This outcome is expected, as the polished transcripts more closely resemble the natural text that LLMs are typically trained on.

\begin{table}[ht!]
    \centering
    \small
    \setlength{\tabcolsep}{4pt}
    \begin{tabular}{l c c | c}
        \toprule
        \multicolumn{4}{c}{\textbf{LLaMA 3.1}} \\
        \midrule
         & \textbf{HCI} & \textbf{HCI} & \\
        & \textbf{EN} & \textbf{DE} & \textbf{Average} \\
        \midrule
        Baseline & 38.70 & 39.40 & 39.05 \\
        RAG w. Slide Text (EN) & 42.40 & \textbf{42.85} & \textbf{42.62} \\ \\
        CPT w. Slide Text (EN) & \underline{\textbf{42.90}} & 33.15 & 38.03 \\
        CPT w. Transcripts (DE) & 40.00 & 33.40 & 36.70 \\
        CPT w. P. Transcripts (DE) & 39.80 & \underline{39.10} & \underline{39.45} \\
        \bottomrule
    \end{tabular}
    \caption[Course-Specific CPT Exam Solving: Performance on HCI Exams]{%
        CPT performance on the HCI course using different course material types. (Exam-level scores, range 0-60.)
    }
    \label{tab:cpt_hci_results}
\end{table}

\section{Conclusion}
This work investigates how university course materials can be leveraged to improve Large Language Models (LLMs) in educational settings. We compare two strategies, Retrieval-Augmented Generation (RAG) and Continual Pre-Training (CPT), to incorporate course-specific knowledge from materials such as lecture slides and transcripts. Our findings show that, given the limited size and multimodal nature of these materials, RAG is a more effective and efficient approach than CPT, which suffers from catastrophic forgetting.  
Furthermore, we demonstrate that incorporating slides as images in a multi-modal RAG setup yields additional performance gains over text-only retrieval, highlighting the importance of preserving visual and structural information in educational resources. Overall, our results suggest that lightweight retrieval-based methods offer a practical path toward developing AI assistants that can better support university-level learning.

\section*{Limitations}
While our results demonstrate the potential of using university course materials to extend LLMs’ scientific knowledge, several limitations remain.
First, our experiments are limited to computer science courses from a single institution. As course structure, teaching style, and content depth vary across universities, the observed improvements may not be the same as for other academic settings.
Second, while we found RAG to be more robust than CPT for small specialized datasets, we did not perform CPT on models other than Llama 3.1 8B, on which CPT might have a more positive effect.
Finally, all evaluations relied on automatic grading using an LLM, which, despite strong correlation with human scores, may introduce systematic biases or grading inconsistencies compared to human assessment.

\section*{Ethical Considerations}
This work uses the publicly available SciEx benchmark and the explicitly authorized university course materials. No personally identifiable information or student data were processed.
Although the goal of this research is to enhance educational accessibility and domain knowledge coverage, there are ethical concerns regarding the potential misuse of LLMs for generating unauthorized teaching material or student assessments. We discourage such uses and emphasize transparent documentation of data sources.

\section*{Acknowledgments}
This work was supported by the Helmholtz Programme-oriented Funding, with project number 46.24.01, project name AI for Language Technologies. 
We acknowledge the HoreKa supercomputer funded by the Ministry of Science, Research and the Arts Baden-Wurttemberg and by the Federal Ministry of Education and Research. This work also received support from the Horizon Europe grant 101213369 DVPS.




\section*{References}\label{sec:reference}

\bibliographystyle{lrec2026-natbib}
\bibliography{both}

\appendix
\section{Prompt Used for Polishing Lecture Transcripts} \label{app:prompt_polish_transcript}
\begin{tcolorbox}[cleanprompt, title=Transcript Polishing Prompt]

You are a professional editor for transcripts of the lecture \texttt{\{lecture\_name\_en\}} with a deep understanding of academic content.

First, you will see the previous, already cleaned section of the transcript. This serves as context to maintain a natural transition:

[Previous section]:

\texttt{\{previous\_synthetic\_chunk\}} or [No previous section available]

Now follows another unedited section of the transcript. Please edit \textbf{only this following section} by:

\begin{itemize}
    \item removing filler words and pauses (e.g., "uh", "so", "exactly", "no", "so to speak"),
    \item correcting grammatical errors,
    \item making the language fluent, precise, and written,
    \item \textbf{removing all sentences} that \textbf{are not part of the technical content of the lecture}. This includes in particular:
    \begin{itemize}
        \item Organizational information (e.g., about homework, exams, slides, Moodle, technical problems),
        \item Comments on the room, time, breaks, sensitivities, or jokes,
        \item Personal anecdotes or conversations with students,
        \item Repetitions without added value, digressions, or small talk.
    \end{itemize}
\end{itemize}

The \textbf{technical content} (e.g., definitions, examples, proofs, concepts) \textbf{must not be summarized or shortened.}

The structure and order of the content should be retained as far as possible.

Please only return the edited, subject-relevant section of text — \textbf{without introduction, explanations, or additional comments}.

\vspace{4pt}
Here is the new section:

\texttt{\{new\_section\}}

\end{tcolorbox}

The prompt is translated into German when polishing German transcripts.

\section{Prompt Used for Generating Exam Answers} \label{app:prompt_answer}

\begin{tcolorbox}[cleanprompt, title=Main Prompt]
    You are a university student. Please answer the following JSON-formatted exam question. \\
    The subquestions (if any) are indexed. \\
    The provided figures (if any) each contains its path at the bottom, which matches the path provided in the JSON. \\

    \texttt{\{context\_message\}} \\

    Please give the answers to the question and subquestions that were asked, and index them accordingly in your output. \\
    You do not have to provide your output in the JSON format. \\
    If you are asked to draw on the figure, then describe with words how you would draw it. \\
    Please provide all answers in English. \\
    Here is the question: \\

    \texttt{\{question\}} or \texttt{\{question\_with\_context\}}
\end{tcolorbox}

The following prompts are used for the RAG experiments.
They are placed in the Main Prompt instead of the \texttt{\{context\_message\}} field.

\begin{tcolorbox}[cleanprompt, title=Context Message: Text Course Material]
    Additionally, you will be provided with some course materials \\
    (can be found in 'Context'). \\
    You can use that additional context to answer the question if it is helpful. \\
    If it is not helpful, you don't have to use it. \\
\end{tcolorbox}

\begin{tcolorbox}[cleanprompt, title=Context Message: Image Course Material]
    You will also be provided with images of course material. \\
    These images each contain a path at the bottom, which corresponds to the path specified in the JSON under 'Context'. \\
    You can use these images to answer the question if this is helpful. \\
    If they are not helpful, you don't have to use them. \\
\end{tcolorbox}

The prompt is translated into German when generating German answers for German questions.

\section{Prompt Used for Grading Exams} \label{app:prompt_grade}
The following prompt is used to grade exams for all experiments.
They were originally used in the SciEx paper \cite{dinh2024sciex}.

\begin{tcolorbox}[cleanprompt, title=Grading Prompts]
    You are a university professor. Please grade the following exam question.

    The exam question, examinee's answer, correct answer and the maximum possible score are provided in the format: \\

    \texttt{[question] <exam\_question> [/question]} \\
    \texttt{[answer] <answer> [/answer]} \\
    \texttt{[correct\_answer] <correct\_answer> [/correct\_answer]} \\
    \texttt{[max\_score] <max\_score> [/max\_score]} \\

    The question is provided in JSON format, but the answer can be freeform text.

    The provided figures in the question (if any) each contain their path at the bottom, which matches the path provided in the JSON.

    The answer is text-only.

    If the question asks to draw on the figure, then the answer should contain a text description of how the drawing should be.

    Please provide the grade between \texttt{[0, <max\_score>]}

    Please provide the reasoning for your grade.

    Please provide your output in the format: \\

    \texttt{[reason] <reasoning> [/reason]} \\
    \texttt{[grade] <grade> [/grade]} \\

    Here is your input: \\

    \texttt{\{question\}, \{answer\}, \{reference\_answer\}, \{max\_points\}}
\end{tcolorbox}

The prompt is translated into German when grading German answers for German questions.

\end{document}